\pdfoutput=1

\documentclass[11pt]{article}

\usepackage[]{EMNLP2022}

\usepackage{times}
\usepackage{latexsym}

\usepackage[T1]{fontenc}

\usepackage[utf8]{inputenc}

\usepackage{microtype}

\usepackage{inconsolata}

\usepackage{graphicx}
\graphicspath{ {figures/} }
\usepackage{caption}
\usepackage{subcaption}
\usepackage{verbatim}
\usepackage{float}
\usepackage{mathtools}

\renewcommand*{\thefootnote}{\fnsymbol{footnote}}

\title{Multi-Tenant Optimization For Few-Shot Task-Oriented FAQ Retrieval}

\author{Asha Vishwanathan \\
  Verloop.io \\
  \texttt{asha@verloop.io} \\\And
  Rajeev Unnikrishnan Warrier\\
  Verloop.io \\
  \texttt{rajeev@verloop.io}\\\AND
  Gautham Vadakkekara Suresh$^{*}$\\
  Verloop.io \\
  \texttt{gautham.suresh09@yahoo.com}\\\And
  Chandra Shekhar Kandpal$^{*}$\\
  Verloop.io \\
  \texttt{chandrashekhar1503@gmail.com}\\}

\begin{document}
\maketitle
\begin{abstract}
\footnotetext{\llap{*}Work done while the authors were working at Verloop.io}
\renewcommand*{\thefootnote}{\arabic{footnote}}
Business-specific Frequently Asked Questions (FAQ) retrieval in task-oriented dialog systems poses unique challenges vis-à-vis community based FAQs. Each FAQ question represents an intent which is usually an umbrella term for many related user queries. We evaluate performance for such Business FAQs both with standard FAQ retrieval techniques using query-Question (q-Q) similarity and few-shot intent detection techniques. Implementing a real-world solution for FAQ retrieval in order to support multiple tenants (FAQ sets) entails optimizing speed, accuracy and cost. We propose a novel approach to scale multi-tenant FAQ applications in real-world context by contrastive fine-tuning of the last layer in sentence Bi-Encoders along with tenant-specific weight switching\footnote{\url{https://github.com/verloop/few-shot-faqir}}.
\end{abstract}

\section{Introduction}

Business-specific Frequently Asked Questions (FAQ) form an important part of many task-oriented dialog systems today. Business FAQs exhibit some common characteristics which differentiate them from community-based FAQs. We show few examples of such FAQs in Table~\ref{tab:QA Examples}. The intent is often a user-defined umbrella term that represents a group of questions in such FAQs. The system needs to respond to user queries which indicate the same intent as that of the FAQ question (Example-1). In some cases, similar FAQ questions can lead to different answers (Example-2). The answers may not always have overlap with the question (Example-3). The response can also be modeled as a system action rather than a text response (Example-4). The system should also be able to identify Out-Of-Scope (OOS) queries and redirect to a human agent.
 
\begin{table*}
\centering
\begin{tabular}{lll}
\hline
\multicolumn{1}{c}{\textbf{No.}} & \multicolumn{1}{c}{\textbf{Question}} & \multicolumn{1}{c}{\textbf{Answer}} \tabularnewline \hline
1 & How to check exchange rate? & Please use the Rate Enquiry option on the Main Menu \tabularnewline \cline{2-2}
 & Rate for Peso & \tabularnewline \hline
2 & Am I going to get a refund & We issue refund for specific products \tabularnewline \cline{2-3}
  & When am I going to get the refund & Refund normally takes 7 days to process \tabularnewline \hline
3 & My transaction failed & Please contact on XXX or mail us at XXX \tabularnewline \cline{2-2}
 & My checkout did not happen properly &  \tabularnewline \hline
4 & Can I speak to an agent? & System action: Transfer to agent \tabularnewline \hline
\end{tabular}
\caption{Examples of actual business questions. All identifying information is masked.}
\label{tab:QA Examples}
\end{table*}

Since each question within an intent category represents it loosely, having a single question for an intent poses a challenge for intent detection. Simple paraphrases do not suffice and therefore, 5-10 variations for each FAQ question representing the intent is sourced from domain experts.

Common approaches in FAQ retrieval include Query-Question similarity and Query-Answer entailment \citep{faq-retrieval-q-q-sim,faq-finder,retrieval-models-q-a-files}. Query-Question (q-Q) methods focus on similarity between a user's query and a question to retrieve the relevant answer. Query-Answer (q-A) methods predict the relevance between the query and an FAQ answer to re-rank the results obtained via q-Q. For FAQ retrieval, we leverage q-Q similarity. The q-A entailment method is infeasible in some FAQ sets due to the generic nature of answers for several different questions, owing to little match between the q-A pair and responses being modeled as system action instead of text. 

We also prefer a single model based approach over stacked models such as retriever-reranker models \citep{zhang-etal-2020-discriminative, gupta-etal-2019-attentive} as inference latency gets compounded in such systems. We compare several q-Q retrieval approaches using pre-trained models and fine-tuned embeddings. We also model the problem as a few-shot intent detection and evaluate classifier based techniques.

From an industry specific dialog solutions perspective, the approaches need to balance accuracy, real time inference latency and costs while having the ability to scale for multiple tenant (FAQ sets) requirements. Our contribution is towards implementing an FAQ retrieval system for Business FAQs in real-world contexts. We describe a low-cost approach to deploy multi-tenant FAQ applications at scale while optimizing for inference latency and accuracy. We also evaluate different methods and models against few-shot intent detection and conversational datasets.
\subsection{Real world constraints}

We process millions of chat messages every day, of which a sizable percentage is related to FAQ retrieval. The expected latency requirement is less than 100 ms under production loads. Tenant specific FAQ sets vary in terms of the domain and intent granularity. It is important to maximize the retrieval accuracy for each tenant while not increasing inference costs drastically.

As on-premise machines are expensive to set-up and maintain, it is often cost-effective and convenient to leverage cloud providers. The application efficiently needs to utilize machine resources and scale for multiple tenants without significant overhead. Since GPU costs are higher, it is cost-effective to train using GPU machines and perform inference with CPU machines.

Apart from this, many tenants require the system to respond with certain standardized answers based on company policy without too many modifications. Therefore, retrieval based solutions are more acceptable than generative solutions. The problems of hallucinations which still exist in generative solutions \citep{roller-etal-2021-recipes, shuster-etal-2021-retrieval-augmentation} constraints the use of these approaches for task-oriented FAQ responses in business contexts.


\section{Related Work}


The models based on transformer architectures have enabled the usage of transfer learning for various tasks such as text similarity \citep{devlin-etal-2019-bert}. \citet{faq-retrieval-q-q-sim} introduced an approach to combine the query-question similarity from TSUBAKI and query-answer relevance from BERT. \citet{arora-etal-2020-hint3} evaluated different providers on three intent recognition datasets, created using queries received by chatbots from real users.

In a few-shot setting, recent approaches leverage the transformer architecture. \citet{casanueva-etal-2020-efficient} demonstrated that using a dual sentence encoder as base model works better in low-resource intent detection tasks as compared to BERT. \citet{zhang-etal-2021-shot} proposed the use of contrastive learning to improve the performance of BERT based classifiers in few-shot intent detection tasks. 

Pre-training methods for creating transferable language representations is a common approach. In-domain pre-training further increases the adaptability of the model to domain related downstream tasks. Specifically for few-shot setting, fine-tuning pre-trained models with supervised learning is found to be effective \citep{zhang-etal-2021-effectiveness-pre}.

Cross-encoder approaches with BERT \citep{devlin-etal-2019-bert} and RoBERTa \citep{roberta} models have achieved SOTA scores using sentence pair training. Training is done with question pairs. During inference, question pairs are created using query and train questions. These pairs are scored by a classifier to predict the final label.

Bi-Encoder models have the advantage of being able to cache the representations and hence, are highly efficient during inference \citep{reimers-gurevych-2019-sentence}. The research on STILTS \citep{stilts} found the advantages of intermediate training on data rich supervised tasks, giving pronounced benefits in data constrained regimes. \citet{casanueva-etal-2020-efficient} discusses the gains achieved via transfer learning using pre-trained sentence encoders as a retriever model.
Cross-Encoder models entail a higher latency as all the question pairs are required to be classified during inference. \citet{zhang-etal-2020-discriminative} explores usage of the classifier model as a re-ranking model following a Bi-Encoder based retriever to reduce inference time.

We evaluate our FAQ retrieval use case with simple baseline approaches, in-domain pre-trained models, classifier based approaches, Cross-Encoders and Bi-Encoders in a few-shot setting.

\section{Datasets}
We choose datasets that are similar to our real-world tenant FAQs and perform experiments on publicly available intent detection datasets for task-oriented dialogues. We use datasets from HINT3 (Curekart, Powerplay11 and SOFMattress) \citep{arora-etal-2020-hint3} which reflects real-world FAQ intents and user queries. We also use DialoGLUE datasets \citep{MehriDialoGLUE2020} - HWU64 \citep{benchmark-conv-agents}, CLINC150 \citep{larson-etal-2019-evaluation} and BANKING77 \citep{casanueva-etal-2020-efficient}.

DialoGLUE datasets further provide train sets for data constrained regimes with 5 and 10 samples per class. Similarly, HINT3 also contains subset variants of datasets which are created by retaining only samples whose entailment score using the ELMo model was less than 0.6. We use the few-shot subsets for DialoGLUE and the full as well as subset variant of HINT3 datasets. These datasets are suitable for few-shot analysis. All the datasets have separate train and test sets and we report our evaluations on the test sets. The CLINC150 and HINT3 datasets have separate in-scope and out-of-scope queries and we use them to evaluate OOS performance. The dataset size, intent and domain distribution is listed in Appendix \ref{sec:appendix}.

\section{Methodology}
\label{sec:methodology}
In this section, we elaborate on our experiments for model training and deployment. We compare different approaches of few-shot FAQ retrieval and describe our deployment strategy for the same.

\subsection{FAQ Retrieval approaches}

\subsubsection{Baseline}
We consider BM25 \citep{mass-etal-2020-unsupervised,faqir-test-collection} and vector space using TF-IDF \citep{faqir-test-collection} as baseline approaches.

\subsubsection{Pre-trained and fine-tuned features}
The following models are considered for feature extraction:
\begin{itemize}
    \item Pre-trained BERT
    \item Pre-trained ConvBERT: ConvBERT model is pre-trained on open-domain conversational data \citep{MehriDialoGLUE2020}.
    \item Fine-tuned ConvBERT: We further fine-tune ConvBERT model as an intent classifier and use the fine-tuned encoder as a feature extractor. 
    \item Pre-trained Bi-Encoders: We use pre-trained Sentence BERT models (\textit{mini-LM-L6-v2, all-mpnet-base-v2}) which are trained on more than 1 Billion sentence pairs consisting of a diverse set of duplicate question pairs, NLI sets, QA pairs.
    \item Fine-tuned Bi-Encoders: We fine-tune pre-trained Bi-Encoders for a Semantic textual similarity task with a contrastive loss \citep{zhang-etal-2021-shot}.
    \item Fine-tuned Task adapted pre-trained Bi-Encoders:  We perform pre-training for adaptation of pre-trained Bi-Encoders with a triplet loss using a cosine distance metric and further fine-tune using a contrastive loss
\end{itemize}
After the features are extracted, inference is implemented using a cosine similarity of query and question embeddings. 

\subsubsection{Classifier}
We fine tune BERT with a linear layer on top to predict the intent class directly \citep{MehriDialoGLUE2020}. 

\subsubsection{Cross-Encoders}
Inspired by the success of Cross-Encoders and the STILTS approach, we fine-tune a pre-trained Cross-Encoder model from SBERT (\textit{stsb-distilroberta-base}). 
\subsection{Question Pair/Triplet Based fine-tuning}
We form Question pairs to fine-tune all Bi-Encoder and Cross-Encoder models. We mark the question pairs belonging to same class as positive samples and ones belonging to different classes as negative samples.

If $c \in C$
Where $C$ is the set of all classes, then we denote $q_{c,i}$ to be the $i^{th}$ element of the $c^{th}$ class. Then, $q_{c,i}\in Q$ where $Q$ is the set of all Questions.
$i \in \{1,\ldots\,n\}$ where $n \in \{5,10,N\}$ when testing for few-shot training with 5 samples, 10 samples or the entire set respectively. 
The question pairs are labelled as defined below.
\begin{align}
\label{eq:query}
L(q_{l,j},q_{m,k}) = \begin{cases} 1 & \text{if $l = m$} \\
0 & \text{if $l \not = m$} \end{cases}
\end{align}
\centerline{\text{$\forall~ l \not = m$ and $j \not = k$}}

This method increases the data size as it generates $^nC_2$ question pairs given $n$ questions. Even in a few-shot setting where there are only 180 samples across 21 intents (SOFMattress), 16110 question pairs were generated.
The Q-Q data can grow exponentially, so the train data is capped to 200K for fine-tuning. We also check fine-tuning on 50K Balanced samples. Hard sampling is done using a probabilistic method based on sample weights. For Q-Q pair with label 0, the weight is equal to the cosine similarity between the Q-Q pair. For label 1, the weight is (1 - cosine similarity) of the Q-Q pair. We use the same pre-trained bi-encoder model for the embeddings which we intend to fine-tune.
Sampling with replacement is then done for labels 0 and 1 to get the required sample sizes.
For the triplet based pre-training, the triplets are constructed such that the anchor and positive belong to the same label and the negative belong to a different label. Q-Q pairs are first constructed using the hard sampling approach described above. For each sample which is the triplet anchor, we get its positives and negatives from the Q-Q pairs.Sampling is then done based on weightage to construct the triplets. 
\subsection{Training and deployment}
The common approach for fine-tuning models involves modifying weights in all layers of the pre-trained model. Instead, we propose an approach where only the final layer of the model is fine-tuned. The weights of all layers which are not fine-tuned are shared across tenant models. During inference in the production environment, we load the base model only once for shared parameters. We keep the tenant-specific weights for all tenants loaded in memory and we switch the tenant specific weights in the model for every inference request. The shared base model weights reduce the memory requirement by a significant margin when we scale to a large number of tenants. Since we retain the tenant weights in memory, replacing the model weights does not result in any significant latency overhead. This allows us to support multiple tenants using the same machine and in-turn reduces the inference costs by an order based on the model size. We select the best model based on expected number of tenants, throughput and memory requirement.
We evaluate the performance of all fine-tuned approaches under the following constraints: Freezing all layers of the encoder except the last layer, training for a fixed number of iterations and using few-shot samples.
\begin{table*}[ht]
\centering
\begin{tabular}{lcccccc}
\hline
\multicolumn{1}{l}{} & \multicolumn{2}{c}{\textbf{BANKING77}} & \multicolumn{2}{c}{\textbf{CLINC150}} & \multicolumn{2}{c}{\textbf{HWU64}} \\
\textbf{Method} & \textbf{5} & \textbf{10} & \textbf{5} & \textbf{10} & \textbf{5} & \textbf{10} \\
\hline
{BM25} & {53.96} & {61.10} & {55.37} & {60.80} & {50.37} & {54.46} \\
{TF-IDF} & {49.51} & {58.14} & {60.91} & {58.14} & {60.91} & {54.36} \\
{BERT - MP} & {40.25} & {48.89} & {68.97} & {66.73} & {52.88} & {57.06} \\
{ConvBERT - MP} & {50.42} & {59.41} & {73.11} & {79.53} & {58.92} & {65.79} \\
{ConvBERT - FT(C)} & {56.98} & {66.91} & {76.44} & {82.91} & {64.86} & {71.46} \\
{SBERT (MiniLM-L6)} & {76.78} & {83.47} & {79.08} & {81.24} & {68.02} & {73.42} \\
{SBERT (MiniLM-L6) - FT(C)} & {80.74} & {86.00} & {84.55} & {87.13} & {76.20} & {79.73} \\
{SBERT (MiniLM-L6) - FT(C) 50K} & {76.42} & {83.18} & {83.97} & {86.86} & {73.97} & {77.60} \\
{SBERT (MiniLM-L6) - FT(T)} & {81.33} & {86.33} & {85.11} & {87.75} & {75.27} & {82.24} \\
{SBERT (MiniLM-L6) - PT-FT(C)} & {84.28} & {84.67} & {89.88} & {89.86} & {85.68} & {86.24} \\
{SBERT (MPNet) - FT(T)} & {83.21} & {\textbf{88.18}} & {88.68} & {91.00} & {78.06} & {83.05} \\
{SBERT (MPNet) - PT-FT(C)} & {\textbf{86.98}} & {87.27} & {\textbf{92.51}} & {\textbf{92.68}} & {\textbf{86.24}} & {\textbf{85.5}} \\
{BERT - FT(C)} & {22.13} & {23.64} & {17.35} & {15.12} & {39.98} & {41.13} \\
{SBERT Cross-Encoder - FT(C)} & {67.10} & {69.83} & {76.10} & {75.20} & {66.91} & {68.03} \\
\hline
\end{tabular}
\caption{Top-1 accuracy of models on the DialoGLUE test sets. MP stands for Mean-Pooling, FT for fine-tuning, (C) for Contrastive loss, (T) for Triplet loss and PT for pre-training. Here, 5 and 10 refers to training subsets created with 5 and 10 samples per intent, respectively.}
\label{tab:exp-results-1}
\end{table*}

\begin{table*}
\centering
\begin{tabular}{lcccccc}
\hline
\multicolumn{1}{l}{} & \multicolumn{2}{c}{\textbf{Curekart}} & \multicolumn{2}{c}{\textbf{Powerplay11}} & \multicolumn{2}{c}{\textbf{SOFMattress}} \\
\textbf{Method} & \textbf{Full Set} & \textbf{Subset} & \textbf{Full Set} & \textbf{Subset} & \textbf{Full Set} & \textbf{Subset} \\
\hline
{BM25} & {72.34} & {71.20} & {51.63} & {49.09} & {58.44} & {52.24}\\
{TF-IDF} & {72.56} & {70.35} & {53.81} & {52.72} & {59.74} & {54.54}\\
{BERT - MP} & {52.87} & {50.21} & {30.54} & {26.18} & {38.09} & {32.03}\\
{ConvBERT - MP} & {65.92} & {63.27} & {42.18} & {37.45} & {49.35} & {47.18}\\
{ConvBERT - FT(C)} & {77.40} & {75.22} & {46.9} & {43.6} & {62.33} & {56.27}\\
{SBERT (MiniLM-L6)} & {82.52} & {79.64} & {64.00} & {62.54} & {74.58} & {71.42}\\
{SBERT (MiniLM-L6) - FT(C)} & {87.38} & {86.06} & {64.00} & {63.2} & {\textbf{78.78}} & {\textbf{77.48}}\\
{SBERT (MiniLM-L6) - FT(C) 50K} & {86.28} & {86.94} & {61.45} & {58.9} & {75.32} & {74.89}\\
{SBERT (MiniLM-L6) - FT(T)} & {86.50} & {85.17} & {61.80} & {\textbf{64.36}} & {77.90} & {74.02}\\
{SBERT (MiniLM-L6) - PT-FT(C)} & {85.39} & {85.39} & {62.54} & {61.81} & {73.16} & {73.59}\\
{SBERT (MPNet) - FT(T)} & {85.61} & {84.51} & {\textbf{66.54}} & {64.00} & {74.45} & {73.59}\\
{SBERT (MPNet) - PT-FT(C)} & {\textbf{88.05}} & {\textbf{87.83}} & {62.18} & {61.81} & {75.32} & {76.19}\\
{BERT - FT(C)} & {55.75} & {58.63} & {18.18} & {18.91} & {41.56} & {38.96}\\
{SBERT Cross-Encoder - FT(C)} & {65.48} & {67.92} & {51.27} & {49.09} & {58.40} & {60.60}\\
\hline
{Dialogflow} & {75.00} & {71.20} & {59.60} & {55.60} & {73.10} & {65.30}\\
{Rasa} & {84.00} & {80.50} & {49.00} & {38.50} & {69.20} & {56.20}\\
{LUIS} & {59.30} & {49.30} & {48.00} & {44.00} & {59.30} & {49.30}\\
{Haptik} & {72.20} & {64.00} & {66.50} & {59.20} & {72.20} & {64.00}\\
{BERT Full Training} & {73.50} & {57.50} & {58.50} & {53.00} & {73.50} & {57.10}\\
\hline
\end{tabular}
\caption{Top-1 accuracy of models on HINT3(v1) test sets with a threshold of 0.1. Here, MP stands for Mean-Pooling, FT for fine-tuning, (C) for Contrastive loss, (T) for Triplet loss and PT for pre-training.}
\label{tab:exp-results-2}
\end{table*}
\section{Experimentation Approaches}
Fine-tuning for all Bi-Encoder and Cross-Encoder models is done using question pairs as elaborated in Section \ref{sec:methodology}. Based on the training method outlined by \citet{mosbach2021on}, we train for a higher number of iterations to offset any fine-tuning instability. For the Bi-Encoders, we fine-tune only the final layer of the model using a contrastive or an online triplet loss. The pre-trained BERT and ConvBERT classifier models were fine-tuned using a softmax cross-entropy loss. The Cross-Encoder was also fine-tuned with the same approach using question pairs with a binary cross-entropy loss. For predicting the query label, we create q-Q pairs limited to 5 questions per intent. The model predicts a score for these combinations, where the label for the question with the highest score is considered to be the predicted label.
We trained all models for 10 K iterations with a learning rate of 2e-5, batch size of 16, AdamW optimizer with 10\% linear warm up and gradient normalization.
\subsection{Pre-training}
Commonly pre-training is done for in-domain adaptation using unlabelled datasets. But in our case, tenant FAQs are spread across multiple niche domains, making it difficult to get domain related data. Moreover, pre-training for each domain separately would result in multiple models per domain, making the hosting costs higher. In case of pre-trained dense retrievers, training on the base language model in an MLM fashion requires a further training of the bi-encoder model. In the GPL paper, \citet{wang2021gpl} adopt a pre-training method which is a triplet based training where the triplets are constructed from the unlabelled data. \citet{gururangan-etal-2020-dont} describe an approach of a second level pre-training using unlabelled corpus and pre-training using task related samples within a domain.

We experiment with a similar approach as GPL but we use the labelled data for the same task available across different domains. We construct triplets using an offline approach instead of an online batch mode, to ensure in-domain triplets. We create 100 K triplets from each dataset and get a total training corpus of 600 K samples. We do a second level of pre-training of the pre-trained bi-encoder models for 140 K iterations, with the Triplet loss using Cosine distance metric and a margin of 0.15. We then use this pre-trained model for further finetuning with a contrastive loss.
\section{Evaluation Metrics}
For evaluation, we consider a combination of FAQ Retrieval metrics along with Intent detection metrics. Similar to the metrics used in \citet{faq-retrieval-q-q-sim}, we consider Top-k accuracy (same as Success Rate referred in the paper), MRR@k (Mean Reciprocal Rank), nDCG@k (normalized Discounted Cumulative Gain) and MAP@k (Mean Average Precision). We also evaluate out-of-scope recall across different thresholds.Responses below a confidence threshold are classified as OOS.
In case the top predicted label score is less than a specified threshold, then the top-k results are shown as suggestions to the user. Metrics like MRR, nDCG and MAP put emphasis on the ordering of the final responses and hence are useful for evaluating the effectiveness of this approach.  
\section{Results}
\begin{figure*}[ht]
     \centering
     \begin{subfigure}[h]{0.49\textwidth}
         \includegraphics[width=75mm]{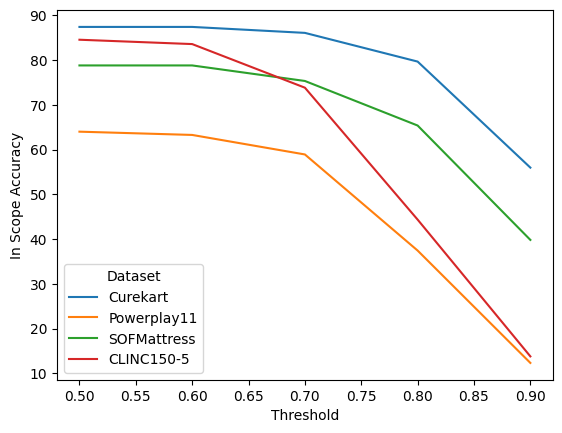}
         \caption{In-scope accuracy}
         \label{fig:in scope accuracy}
     \end{subfigure}
     ~
     \begin{subfigure}[h]{0.49\textwidth}
         \includegraphics[width=75mm]{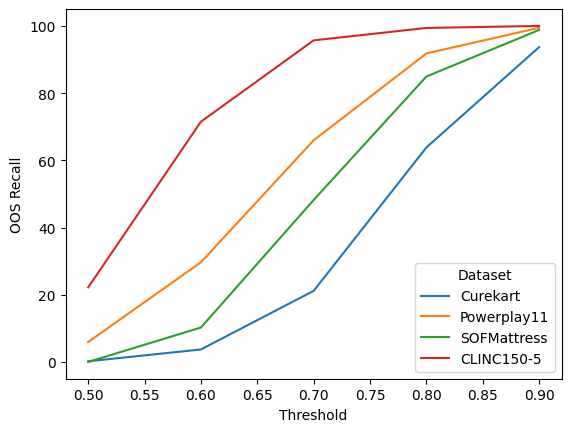}
         \caption{OOS recall}
         \label{fig:oos recall}
     \end{subfigure}
        \caption{In-scope accuracy and OOS recall on different datasets for fine-tuned MiniLM-L6 model.}
        \label{fig:oos-details}
\end{figure*}

Table~\ref{tab:exp-results-1} and \ref{tab:exp-results-2} show the Top-1 accuracy for all experiments. 
Table~\ref{tab:exp-results-3} shows the performance of the MiniLM-L6 model in top-3 setting. Figure~\ref{fig:oos-details} further shows its OOS recall and in-scope accuracy against different thresholds.


From our experiments, we find that the fine-tuned Bi-Encoder models perform the best across all datasets. All the Bi-Encoder embedding approaches significantly outperform the other approaches in few-shot setting in spite of being constrained with the last layer fine-tuning strategy.

For HINT3 datasets, we use the benchmarking results as reported by \citet{arora-etal-2020-hint3} for comparison. We find that the fine-tuned Bi-Encoder shows better performance than all benchmarked chatbot solutions on both full and constrained datasets. 
 We noticed improvement while fine-tuning the pre-trained Bi-Encoder models using the Question pair approach. We also note that online triplet loss based fine-tuning performed marginally better as compared to contrastive Loss for all the base models other than MPNet  model on HINT3 dataset. We observe that the question pair / triplet training strategy mitigates the effect of fewer samples. Reducing sample sizes of the Q-Q pairs from 200 K to 50 K balanced samples hurts the performance across all datasets.

Fine-tuning the task adapted pre-trained models shows much better results in comparison to the base pre-trained models in most of the Dialoglue datasets but the gains are either less or deteriorate for HINT3 datasets. In HINT3 datasets, the median examples per intent is quite low as compared to Dialoglue, with some intents having only 1 sample. There would be no triplets formed for such samples. We also see a wide variation in the number of examples per intent ranging from 3 to 95 in the Curekart dataset. This would reduce the number of triplets for such intents. 

While Cross-Encoder strategies are supposed to be superior, we find that the Cross-Encoder fine-tuning suffers due to the restrictive training strategy and performs sub-par when compared to the Bi-Encoders.
 
The BERT based classifier also shows poor performance which appears to degrade on datasets with more classes. In comparison to training BERT model with all layers unfreezed as reported in the \citet{arora-etal-2020-hint3}, we see that the last layer fine-tuning strategy severely impacts the model performance.

We also observe that models pre-trained on in-domain data such as ConvBERT is superior to base BERT even as a feature extractor. Interestingly, we see that even baseline approaches with BM25 and TF-IDF show better results than pre-trained BERT in such settings. Fine-tuning the ConvBERT model on the supervised classification task shows an improvement over the pre-trained model. If we look at the OOS recall for the fine-tuned MiniLM-L6 model, the similarity scores are normally high and lower thresholds do not have much impact. We further observe that each dataset has different thresholds where the best trade-off between OOS recall and in-scope accuracy is achieved. The top-3 accuracy of MiniLM-L6 on all datasets including the challenging Powerplay11 dataset is above 70\%. We choose the fine-tuned Bi-Encoder models based on superior performance with our training strategy.

\begin{table*}
\centering
\begin{tabular}{lcccccccc}
\hline
\multicolumn{1}{l}{} & \multicolumn{2}{c}{\textbf{Curekart}} & \multicolumn{2}{c}{\textbf{Powerplay11}} & \multicolumn{2}{c}{\textbf{SOFMattress}} & \multicolumn{2}{c}{\textbf{CLINC150}} \\
\textbf{Metrics} & \textbf{Full} & \textbf{Subset} & \textbf{Full} & \textbf{Subset} & \textbf{Full} & \textbf{Subset} & \textbf{Sample-5} & \textbf{Sample-10} \\
\hline
{Success Rate} & {89.80} & {88.93} & {73.81} & {73.09} & {81.38} & {81.81} & {94.22} & {95.08}\\
{MRR} & {88.38} & {87.38} & {68.42} & {67.45} & {79.94} & {79.43} & {88.88} & {90.77}\\
{nDCG} & {89.33} & {88.69} & {72.74} & {71.48} & {81.06} & {81.33} & {93.09} & {94.33}\\
{MAP} & {88} & {87} & {68} & {67} & {80} & {79} & {89} & {91}\\
\hline
\end{tabular}
\caption{Top-3 accuracy of fine-tuned MiniLM-L6 on the OOS datasets}
\label{tab:exp-results-3}
\end{table*}
\section{Deployment}
\label{sec:deployment}
\begin{table}[hb]
\centering
\begin{tabular}{lcccc}
\hline
\textbf{Model} & \textbf{Median} & \textbf{90\%ile} & \textbf{RPS}\\
\textbf{} & \textbf{latency} & \textbf{latency} & \textbf{}\\
\hline
{MiniLM-L6} & {93} & {260} & {22.9} \\
{MPNet} & {210} & {500} & {6.2} \\
\hline
\end{tabular}
\caption{Load testing results with latency and requests per second (RPS) for best models. All latency values in milliseconds (ms).}
\label{tab:locust-results}
\end{table}
\begin{table}
\centering
\begin{tabular}{lcc}
\hline
\multicolumn{1}{l}{\textbf{Model}} & \multicolumn{2}{c}{\textbf{Memory requirement}}\\
\textbf{} & \textbf{Full model} & \textbf{Weight switch}\\
\hline
{MiniLM-L6} & {110} & {20.5}\\
{MPNet} & {560} & {41.5}\\
\hline
\end{tabular}
\caption{Memory increase per tenant for the best performing models. All values are in MegaBytes (MBs)}
\label{tab:memory-results}
\end{table}
Smaller models give us  more scalability at the cost of accuracy while larger models tend to have lower throughput and higher memory requirements. The model selection is done based on the specific business need which dictate these three metrics.
Table~\ref{tab:locust-results} shows the load testing results using Locust framework, where we measure the mean latency per request and the request per second (RPS) achieved. The models were tested under increasing loads until RPS was stagnant and latency shot up. The results depict the performance at the maximum load the model can comfortably handle. The models are hosted via Kubernetes and deployed as a microservice using native Pytorch inference. Machine configuration used for locust testing was c2-standard-4 machine (4 vCPU and 16GB RAM) on Google Cloud Platform. From Table~\ref{tab:memory-results}, we find that weight switching saves 81.5\% memory per tenant for MiniLM-L6. For the larger model MPNet, it saves 92.5\% per tenant. The benefits of weight switching increases as model sizes gets larger. 
\setcounter{footnote}{1}
\renewcommand*{\thefootnote}{\arabic{footnote}}
We see that fine-tuned Bi-Encoder models work best in terms of accuracy in few-shot setting. We also have a latency benefit as the question embeddings are computed upfront and cached. At inference, prediction are done as cosine similarity between the query embeddings and the cached question embeddings. For further gains, we cache these tenant specific embeddings using FAISS or ANNOY
\footnotemark. While the task adapted pre-trained models worked quite well in terms of accuracy, it entails repeated pre-training of the model for every new tenant from a niche domain and the subsequent fine-tuning. It poses a hindrance in decoupling tenant data seamlessly. However, it is still a very viable approach in case of separate hosting of domain specific models.
We choose MiniLM-L6 as it handles higher load at significantly lower latency as compared to the best performing model MPNet, with a slight trade-off in accuracy.\footnotetext{\url{https://github.com/spotify/annoy}}
\section{Conclusion}
We evaluated various methods for retrieval of Business FAQs by modeling the problem as an FAQ retrieval and a few-shot intent detection task. We proposed a realistic multi-tenant deployment solution with trade-offs in accuracy while balancing for cost and latency. Our last layer weight-switching strategy works well where the model has to be fine-tuned on tenant specific tasks. We observed that fine-tuned Bi-Encoder embeddings work best in a few-shot setting even under a constrained fine-tuning strategy.

\section{Limitations}
Although the weight switching approach is highly scalable with increase in number of tenants, it has few limitations. Our last-layer weight switching strategy is more effective for heavier models and the benefit starts to diminish with lighter models. Although this approach is highly scalable, implementing it for inference frameworks such as ONNX Runtime and NVIDIA Triton is not straightforward. Hence, we are currently limited to using native PyTorch inference which has higher latency. 
Apart from this, fine-tuning only the last layer also constraints the model training leading to lower accuracy as compared to training all the layers. All our experiments showcase benefits only with base models and do not showcase inference benefits using other strategies such as quantization, pruning, ANNs and caching. Such strategies can make the system more scalable.

\section{Acknowledgements}
This work was supported by Verloop.io. We wish to thank Gaurav Singh and Peeyush Jain from Verloop. We thank the anonymous reviewers whose thoughtful comments improved our final work.

\bibliography{anthology,faqir}
\bibliographystyle{acl_natbib}
\appendix
\section*{Appendix}
\section{Dataset details}
\label{sec:appendix}
The details regarding different datasets is given in Table~\ref{tab:datasizes-7}. The dataset BANKING77 consists of 77 intents for 1 domain, CLINC150 has 150 intents across 10 domains and HWU64 has 64 intents across 21 domains. Curekart, Powerplay11 and SOFMattress have 28, 59 and 21 intents respectively. Powerplay11 is a challenging set with some intents having only 1 sample even in full dataset and median samples per intent being 3 in the subset variant. The total number of Query-Question (Q-Q) pairs generated for each dataset is also shown.
\begin{table*}
\centering
\begin{tabular}{lcccccc}
\hline
\textbf{Datasets} & \textbf{Intents (Domains)} & \textbf{Min} & \textbf{Max} & \textbf{Median} & \textbf{Samples} & \textbf{q-Q Pairs} \\
\hline
{BANKING77} & {77 (1)} & {30} & {167} & {108} & {8.6K} & {-}\\
{BANKING77-5} & {77 (1)} & {5} & {5} & {5} & { 385} & {73920}\\
{BANKING77-10} & { 77 (1)} & {10} & {10} & {10} & { 770} & {296065}\\
{CLINC150} & {150 (10)} & {100} & {100} & {100} & {15K} & {-}\\
{CLINC150-5} & {150 (10)} & {5} & {5} & {5} & { 750} & {280875}\\
{CLINC150-10} & {150 (10)} & {10} & {10} & {10} & { 1500} & {1124250}\\
{HWU64} & {64 (21)} & {33} & {159} & {156} & { 8.9K} & {-}\\
{HWU64-5} & {64 (21)} & {5} & {5} & {5} & { 320} & {51040}\\
{HWU64-10} & { 64 (21)} & {10} & {10} & {10} & {640} & {204480}\\
{Curekart} & {28 (1)} & {3} & {95} & {14} & {600} & {179700}\\
{Curekart-Subset} & {28 (1)} & {2} & {72} & {8} & {413} & {85078}\\
{SOFmattress} & {21 (1)} & {9} & {34} & {12} & {328} & {53628}\\
{SOFmattress-Subset} & {21 (1)} & {3} & {18} & {7} & {180} & {16110}\\
{Powerplay} & {59 (1)} & {1} & {46} & {7} & { 471} & {110685}\\
{Powerplay-Subset} & {59 (1)} & {1} & {24} & {3} & {261} & {33930}\\
\hline
\end{tabular}
\caption{Details of datasets used along with number of Q-Q pair generated.Min,Max and Median are on samples per intent}
\label{tab:datasizes-7}
\end{table*}

\end{document}